# Expectation propagation for approximate inference in dynamic Bayesian networks

**Tom Heskes**    **Onno Zoeter**
SNN, University of Nijmegen
Geert Grooteplein 21, 6252 EZ, Nijmegen, The Netherlands
{*tom, orzoeter*}*@snn.kun.nl*

## Abstract

We describe expectation propagation for approximate inference in dynamic Bayesian networks as a natural extension of Pearl's exact belief propagation. Expectation propagation is a greedy algorithm, converges in many practical cases, but not always. We derive a double-loop algorithm, guaranteed to converge to a local minimum of a Bethe free energy. Furthermore, we show that stable fixed points of (damped) expectation propagation correspond to local minima of this free energy, but that the converse need not be the case. We illustrate the algorithms by applying them to switching linear dynamical systems and discuss implications for approximate inference in general Bayesian networks.

## 1  INTRODUCTION

Algorithms for approximate inference in dynamic Bayesian networks can be roughly divided into two categories: sampling approaches and variational approaches. Popular sampling approaches in the context of dynamic Bayesian networks are so-called particle filters. Examples of variational approaches for dynamic Bayesian algorithms are (Ghahramani and Hinton, 1998) for switching linear dynamical systems and (Ghahramani and Jordan, 1997) for factorial hidden Markov models. A subset of the variational approaches are methods based on greedy projection. These are similar to standard belief propagation, but include a projection step to a simpler approximate belief. Examples are the extended Kalman filter, generalized pseudo-Bayes for switching linear dynamical systems (Bar-Shalom and Li, 1993), and the Boyen-Koller algorithm for hidden Markov models (Boyen and Koller, 1998). In this article, we will focus on these greedy projection algorithms.

Expectation propagation (Minka, 2001b) stands for a whole family of approximate inference algorithms that includes loopy belief propagation (Murphy et al., 1999) and many (improved and iterative versions of) greedy projection algorithms as special cases. In Section 2 we will describe expectation propagation in dynamic Bayesian networks as an extension of exact belief propagation, the only difference being an additional projection (collapse) in the procedure for updating messages. We illustrate the resulting procedure in Section 2.6 on switching linear dynamical systems.

Expectation propagation does not always converge (Minka, 2001a). In Section 3 we therefore derive a double-loop algorithm that guarantees convergence to a minimum of a Bethe free energy. Rephrasing the optimization as a saddle-point problem, we can interpret damped expectation propagation as an attempt to perform gradient descent-ascent.

Simulation results regarding approximate belief propagation applied to switching linear dynamical systems are presented in Section 4. In Section 5 we end with conclusions and a discussion of implications for approximate inference in general Bayesian networks.

## 2  EXPECTATION PROPAGATION AS COLLAPSE-PRODUCT RULE

### 2.1  DYNAMIC BAYESIAN NETWORKS

We consider general dynamic Bayesian networks with latent variables $\mathbf{x}_t$ and observations $\mathbf{y}_t$. The graphical model is visualized in Figure 1 for $T = 4$ time slices. The joint distribution of latent variables $\mathbf{x}_{1:T}$ and observables $\mathbf{y}_{1:T}$ can be written in the form

$$P(\mathbf{x}_{1:T}, \mathbf{y}_{1:T}) = \prod_{t=1}^{T} \psi_t(\mathbf{x}_{t-1}, \mathbf{x}_t, \mathbf{y}_t),$$

where

$$\psi_t(\mathbf{x}_{t-1}, \mathbf{x}_t, \mathbf{y}_t) = P(\mathbf{x}_t|\mathbf{x}_{t-1})P(\mathbf{y}_t|\mathbf{x}_t),$$

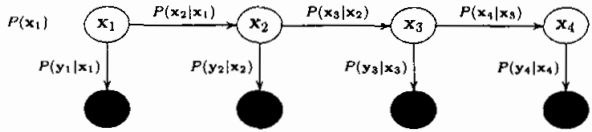

Figure 1: Graphical representation of a dynamic Bayesian network.

and with the convention $\psi_1(\mathbf{x}_0, \mathbf{x}_1, \mathbf{y}_1) \equiv \psi_1(\mathbf{x}_1, \mathbf{y}_1)$, i.e., $P(\mathbf{x}_1|\mathbf{x}_0) = P(\mathbf{x}_1)$, the prior. In the following we will not pay special attention to the boundaries: details can be worked out easily. We will assume that all evidence $\mathbf{y}_{1:T}$ is fixed and given and include it in the definition of the potentials $\psi_t(\mathbf{x}_{t-1,t}) \equiv \psi_t(\mathbf{x}_{t-1}, \mathbf{x}_t, \mathbf{y}_t)$. $\mathbf{x}_t$ can be thought of as a "supernode" containing all latent variables for time-slice $t$, which can include both discrete and continuous variables (as e.g. in switching linear dynamical systems). For convenience we will stick to integral notation.

### 2.2 THE COLLAPSE-PRODUCT RULE

Our goal is to compute one-slice marginals or "beliefs" of the form $P(\mathbf{x}_t|\mathbf{y}_{1:T})$: the probability of the latent variables in a time slice given all evidence. This marginal is required in many EM-type learning procedures, but can also be of interest by itself, especially when the latent variables have a direct interpretation. Two-slice marginals $P(\mathbf{x}_{t-1,t}|\mathbf{y}_{1:T})$ and the data likelihood $P(\mathbf{y}_{1:T})$ are then obtained more or less for free.

A well-known procedure for computing beliefs in general Bayesian networks is Pearl's belief propagation (Pearl, 1988). Here we will follow a description of belief propagation as a specific case of the sum-product rule in factor graphs (Kschischang et al., 2001). This description is symmetric with respect to the forward and backward messages. We distinguish variable nodes $\mathbf{x}_t$ and local function nodes $\psi_t$ in between variable nodes $\mathbf{x}_{t-1}$ and $\mathbf{x}_t$. The message from $\psi_t$ forward to $\mathbf{x}_t$ is called $\alpha_t(\mathbf{x}_t)$ and the message from $\psi_t$ back to $\mathbf{x}_{t-1}$ is referred to as $\beta_{t-1}(\mathbf{x}_{t-1})$ (see Figure 2).

The belief at variable node $\mathbf{x}_t$ is the product of all messages sent from neighboring local function nodes:

$$P(\mathbf{x}_t|\mathbf{y}_{1:T}) \propto \alpha_t(\mathbf{x}_t)\beta_t(\mathbf{x}_t) .$$

The sum-product rule for factor graphs implies that in a chain, variable nodes simply pass the messages that they receive on to the next local function node.

Information about the potentials is incorporated at the corresponding local function nodes. We extend the standard recipe for computing the message from the local function node $\psi_t$ to a neighboring variable node $\mathbf{x}_{t'}$, where $t'$ can be either $t$ (forward message) or $t-1$ (backward message), as follows.

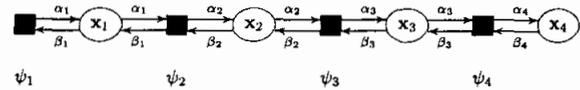

Figure 2: Message propagation.

1. Multiply the potential corresponding to the local function node $\psi_t$ with *all* messages from neighboring variable nodes to $\psi_t$, yielding

$$\hat{P}(\mathbf{x}_{t-1,t}) \propto \alpha_{t-1}(\mathbf{x}_{t-1})\psi_t(\mathbf{x}_{t-1,t})\beta_t(\mathbf{x}_t) , \quad (1)$$

our current estimate of the distribution at the local function node given the incoming messages $\alpha_t(\mathbf{x}_{t-1})$ and $\beta_t(\mathbf{x}_t)$.

2. Integrate out all variables except variable $\mathbf{x}_{t'}$ to obtain the current estimate of the belief state $\hat{P}(\mathbf{x}_{t'})$ *and project (collapse) this belief state onto a distribution in the exponential family, yielding the approximate belief* $q_{t'}(\mathbf{x}_{t'})$.

3. Conditionalize, i.e., divide by the message from $\mathbf{x}_{t'}$ to $\psi_t$.

Without the collapse operation in step 2, we obtain the standard sum-product rule in a slight disguise. The usual definition *excludes* in step 1 the incoming message from $\mathbf{x}_{t'}$ to $\psi_t$ and has no division afterwards. However, since without collapse "multiplication + marginalization + division = marginalization", this essentially gives the same procedure. With collapse, the ordering does matter: "multiplication + collapse + division $\neq$ collapse". An important lesson of expectation propagation, which is repeated here, is that it makes better sense to approximate beliefs and derive the messages from these approximate beliefs than to approximate the messages themselves.

### 2.3 THE EXPONENTIAL FAMILY

For the approximating family of distributions we take a particular member of the exponential family, i.e.,

$$q_t(\mathbf{x}_t) \propto e^{\gamma_t^T \mathbf{f}(\mathbf{x}_t)} , \quad (2)$$

with $\gamma_t$ the canonical parameters and $\mathbf{f}(\mathbf{x}_t)$ the sufficient statistics. Typically, $\gamma$ and $\mathbf{f}(\mathbf{x})$ are vectors with many components. Examples are Gaussian, Poisson, Wishart, multinomial, Boltzmann, and conditional Gaussian distributions, among many others.

If we initialize the forward and backward messages as

$$\alpha_t(\mathbf{x}_t) \propto e^{\alpha_t^T \mathbf{f}(\mathbf{x}_t)} \text{ and } \beta_t(\mathbf{x}_t) \propto e^{\beta_t^T \mathbf{f}(\mathbf{x}_t)} ,$$

for example choosing $\alpha_t = \beta_t = 0$, they will stay of this form: the canonical parameters $\alpha_t$ and $\beta_t$ fully specify the messages $\alpha_t(\mathbf{x}_t)$ and $\beta_t(\mathbf{x}_t)$ and are all that we have to keep track of. As in exact belief propagation, the belief $q_t(\mathbf{x}_t)$ is defined as the product of incoming messages, i.e., is of the form (2) with $\gamma_t = \alpha_t + \beta_t$.

Typically, there are two kinds of reasons for making a particular choice within the exponential family. Both can be treated within this same framework.

- The exact belief is not in the exponential family and therefore difficult to handle. The approximating distribution is of a particular exponential form, but usually further completely free. Examples are a Gaussian for the nonlinear Kalman filter or a conditional Gaussian for the switching linear dynamical system treated in Section 2.6.

- The exact belief is in the exponential family, but requires too many variables to fully specify it. The approximate belief is part of the same exponential family but with additional constraints, e.g., factorized over (groups of) variables as in the Boyen-Koller algorithm (Boyen and Koller, 1998).

### 2.4 MOMENT MATCHING

In the projection step, we replace the current estimate $\hat{P}(\mathbf{x})$ by the approximate $q(\mathbf{x})$ of the form (2) closest to $\hat{P}(\mathbf{x})$ in terms of the Kullback-Leibler divergence

$$\mathrm{KL}(\hat{P}|q) = \int d\mathbf{x}\, \hat{P}(\mathbf{x}) \log\left[\frac{\hat{P}(\mathbf{x})}{q(\mathbf{x})}\right].$$

With $q(\mathbf{x})$ in the exponential family, the solution follows from moment matching: we have to find the canonical parameters $\boldsymbol{\gamma}$ such that

$$\mathbf{g}(\boldsymbol{\gamma}) \equiv \langle \mathbf{f}(\mathbf{x})\rangle_q \equiv \int d\mathbf{x}\, q(\mathbf{x})\mathbf{f}(\mathbf{x}) = \int d\mathbf{x}\, \hat{P}(\mathbf{x})\mathbf{f}(\mathbf{x}).$$

For members of the exponential family the link function $\mathbf{g}(\boldsymbol{\gamma})$ is unique and invertible: there is a one-to-one mapping from canonical parameters to moments.

### 2.5 FORWARD AND BACKWARD

Working out the moment matching (step 2) and division (step 3) in terms of the canonical parameters $\alpha_t$ and $\beta_t$ and the two-slice marginals $\hat{p}_t(\mathbf{x}_{t-1,t}) \equiv \hat{P}(\mathbf{x}_{t-1,t})$ of (1), we arrive at the following forward and backward passes.

**Forward pass.** Compute $\alpha_t$ such that

$$\langle \mathbf{f}(\mathbf{x}_t)\rangle_{\hat{p}_t} = \langle \mathbf{f}(\mathbf{x}_t)\rangle_{q_t} = \mathbf{g}(\alpha_t + \beta_t). \quad (3)$$

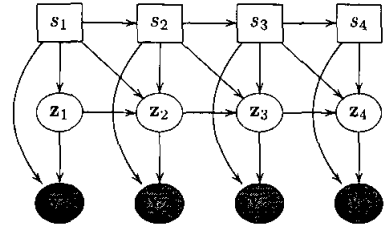

Figure 3: Switching linear dynamical system.

Note that $\langle \mathbf{f}(\mathbf{x}_t)\rangle_{\hat{p}_t}$ only depends on the messages $\alpha_{t-1}$ and $\beta_t$. With $\beta_t$ kept fixed, the solution $\alpha_t = \tilde{\alpha}_t(\alpha_{t-1}, \beta_t)$ can be computed by inverting $\mathbf{g}(\cdot)$, i.e., translating from a moment form to a canonical form: $\alpha_t = g^{-1}(\langle \mathbf{f}(\mathbf{x}_t)\rangle_{\hat{p}_t}) - \beta_t$.

**Backward pass.** Compute $\beta_{t-1}$ such that

$$\langle \mathbf{f}(\mathbf{x}_{t-1})\rangle_{\hat{p}_t} = \langle \mathbf{f}(\mathbf{x}_{t-1})\rangle_{q_{t-1}} = \mathbf{g}(\alpha_{t-1} + \beta_{t-1}). \quad (4)$$

Similar to the forward pass, the solution can be written $\beta_{t-1} = \tilde{\beta}_{t-1}(\alpha_{t-1}, \beta_t)$.

The order in which the messages are updated is free to choose. However, iterating the standard forward-backward passes seems to be most logical.

Without collapse, i.e., if the exponential family distribution is not an approximation but exact, we have a standard forward-backward algorithm. In these cases, one can easily show that $\tilde{\alpha}_t(\alpha_{t-1}, \beta_t) = \tilde{\alpha}_t(\alpha_{t-1})$, independent of $\beta_t$ and similarly $\tilde{\beta}_{t-1}(\alpha_{t-1}, \beta_t) = \tilde{\beta}_{t-1}(\beta_t)$: the forward and backward messages do not interfere and there is no need to iterate. This is the case for the two-filter version of the Kalman smoother and for the forward-backward algorithm for hidden Markov models.

### 2.6 EXAMPLE: SWITCHING LINEAR DYNAMICAL SYSTEM

Here we will illustrate the operations required for application of expectation propagation to switching linear dynamical systems. Reliable algorithms for approximate inference are very relevant, since exact inference in switching linear dynamical systems is NP-hard (Lerner and Parr, 2001): the number of mixture components needed to describe the exact distribution grows exponentially over time.

The potential corresponding to the switching linear dynamical system graphically visualized in Figure 3 can be written

$$\psi_t(s_{t-1,t}^{i,j}, \mathbf{z}_{t-1,t}) = \\ p_\psi(s_t^j|s_{t-1}^i)\Phi(\mathbf{z}_t; A_{ij}\mathbf{z}_{t-1}, Q_{ij})\Phi(\mathbf{y}_t; C_j\mathbf{z}_t, R_j),$$

where $\Phi(\mathbf{z}; \mathbf{m}, V)$ stands for a Gaussian with mean $\mathbf{m}$ and covariance matrix $V$ and with shorthand notation $s_t^i$ for $s_t = i$ and $s_{t-1,t}^{i,j}$ for $s_{t-1} = i$ and $s_t = j$. The messages are taken to be conditional Gaussian potentials of the form

$$\alpha_{t-1}(s_{t-1}^i, \mathbf{z}_{t-1}) \propto p_\alpha(s_{t-1}^i) \Psi(\mathbf{z}_{t-1}; \mathbf{m}_{i,t-1}^\alpha, V_{i,t-1}^\alpha)$$
$$\beta_t(s_t^j, \mathbf{z}_t) \propto p_\beta(s_t^j) \Psi(\mathbf{z}_t; \mathbf{m}_{j,t}^\beta, V_{j,t}^\beta),$$

where the potential $\Psi(\mathbf{z}; \mathbf{m}, V)$ is of the same form as $\Phi(\mathbf{z}; \mathbf{m}, V)$, but without the normalization and in fact need not be normalizable, i.e., can have a negative covariance. The message $\alpha_{t-1}(s_{t-1}, \mathbf{z}_{t-1})$ is a combination of $M$ Gaussian potentials, one for each switch state $i$, and can always be written in exponential form. The two-slice marginal of (1),

$$\hat{P}(s_{t-1,t}^{i,j}, \mathbf{z}_{t-1,t}) \propto$$
$$\alpha_{t-1}(s_{t-1}^i, \mathbf{z}_{t-1}) \psi_t(s_{t-1,t}^{i,j}, \mathbf{z}_{t-1,t}) \beta_t(s_t^j, \mathbf{z}_t),$$

consists of $M^2$ Gaussians: one for each $\{i,j\}$. With some bookkeeping, which involves the translation from canonical parameters to moments, we can rewrite

$$\hat{P}(s_{t-1,t}^{i,j}, \mathbf{z}_{t-1,t}) \propto \hat{p}_{ij} \Phi(\mathbf{z}_{t-1,t}; \hat{\mathbf{m}}_{ij}, \hat{V}_{ij}), \quad (5)$$

where $\hat{\mathbf{m}}_{ij}$ is a $2N$-dimensional vector and $\hat{V}_{ij}$ a $2N \times 2N$ covariance matrix.

In the forward pass (3), we have to compute the moments of $\hat{P}(s_{t-1,t}, \mathbf{z}_{t-1,t})$ that follow by integrating out $\mathbf{z}_{t-1}$ and summing over $s_{t-1}$. The integration over $\mathbf{z}_{t-1}$ can be done exactly:

$$\hat{P}(s_{t-1,t}^{i,j}, \mathbf{z}_t) \propto \hat{p}_{ij} \Phi(\mathbf{z}_t; \hat{\mathbf{m}}_{ij}, \hat{V}_{ij}),$$

where now $\hat{\mathbf{m}}_{ij}$ and $\hat{V}_{ij}$ are supposed to be restricted to the components corresponding to $\mathbf{z}_t$, i.e., the components $N+1$ to $2N$ in the means and covariances of (5). Summation over $s_{t-1}$ yields a mixture of $M$ Gaussians for each switch state $j$, which is *not* a member of the exponential family. The conditional Gaussian

$$q_t(s_t^j, \mathbf{z}_t) = \hat{p}_j \Phi(\mathbf{z}_t; \hat{\mathbf{m}}_j, \hat{V}_j)$$

closest in KL-divergence to this mixture of Gaussians follows from moment matching:

$$\hat{p}_j = \sum_i \hat{p}_{ij}, \quad \hat{\mathbf{m}}_j = \sum_i \hat{p}_{i|j} \hat{\mathbf{m}}_{ij}, \quad \text{and}$$
$$\hat{V}_j = \sum_i \hat{p}_{i|j} \hat{V}_{ij} + \sum_i \hat{p}_{i|j} (\hat{\mathbf{m}}_{ij} - \hat{\mathbf{m}}_i)(\hat{\mathbf{m}}_{ij} - \hat{\mathbf{m}}_i)^T,$$

with $p_{i|j} = p_{ij}/p_j$. To find the new forward message $\alpha_t(s_t, \mathbf{z}_t)$ we have to divide the approximate belief $q_t(s_t, \mathbf{z}_t)$ by the backward message $\beta_t(s_t, \mathbf{z}_t)$. This is most easily done by translating $q_t(s_t, \mathbf{z}_t)$ from the moment form above to a canonical form and subtracting the canonical parameters corresponding to $\beta_t(s_t, \mathbf{z}_t)$ to yield the new $\alpha_t(s_t, \mathbf{z}_t)$ in canonical form.

The procedure for the backward pass (4) follows in exactly the same manner by integrating out $\mathbf{z}_t$ and collapsing over $s_t$. The forward filtering pass is equivalent to a method called GPB2 (Bar-Shalom and Li, 1993), one of the current most popular inference algorithms for a switching linear dynamical system. An attempt has been made to obtain a similar smoothing procedure, but this required quite some additional approximations (Murphy, 1998). In the above description however, forward and backward passes are completely symmetric and smoothing does not require any approximations beyond the ones already made for filtering. Furthermore, the forward and backward passes can be iterated until convergence to find a consistent and better approximation. In a similar way, one can apply expectation propagation to iteratively improve other approximate methods for inference in dynamic Bayesian networks. An iterative version of the Boyen-Koller algorithm (Boyen and Koller, 1998) has been proposed in (Murphy and Weiss, 2001).

## 3 OPTIMIZING A FREE ENERGY

### 3.1 THE FREE ENERGY

Fixed points of expectation propagation correspond to fixed points of the "Bethe free energy" (Minka, 2001b)

$$F(\hat{p}, q) = -\sum_{t=1}^{T-1} \int d\mathbf{x}_t \, q_t(\mathbf{x}_t) \log q_t(\mathbf{x}_t)$$
$$+ \sum_{t=1}^{T} \int d\mathbf{x}_{t-1,t} \, \hat{p}_t(\mathbf{x}_{t-1,t}) \log \left[ \frac{\hat{p}_t(\mathbf{x}_{t-1,t})}{\psi_t(\mathbf{x}_{t-1,t})} \right], \quad (6)$$

under expectation constraints

$$\langle \mathbf{f}(\mathbf{x}_t) \rangle_{\hat{p}_t} = \langle \mathbf{f}(\mathbf{x}_t) \rangle_{q_t} = \langle \mathbf{f}(\mathbf{x}_t) \rangle_{\hat{p}_{t+1}}. \quad (7)$$

Here $\hat{p}$ refers to all two-slice marginals and $q$ to all one-slice marginals, by definition of the exponential form (2). This free energy is equivalent to the Bethe free energy for (loopy) belief propagation in (Yedidia et al., 2001), with the stronger marginalization constraints replaced by the weaker expectation constraints (7) that correspond to the projection step in the collapse-product rule. We are specifically interested in minima of this free energy.

### 3.2 A DOUBLE-LOOP ALGORITHM

The technical problem is that the free energy (6) may not be convex in $\{\hat{p}, q\}$ under the constraints (7), especially because of the concave $-q \log q$-term. Bounding this concave part linearly, we obtain

$$F_{\text{bound}}(\hat{p}, q, q^{\text{old}}) = -\sum_{t=1}^{T-1} \int d\mathbf{x}_t \, q_t(\mathbf{x}_t) \log q_t^{\text{old}}(\mathbf{x}_t)$$
$$+ \sum_{t=1}^{T} \int d\mathbf{x}_{t-1,t} \, \hat{p}_t(\mathbf{x}_{t-1,t}) \log \left[ \frac{\hat{p}_t(\mathbf{x}_{t-1,t})}{\psi_t(\mathbf{x}_{t-1,t})} \right] . \quad (8)$$

This formulation suggests a double-loop algorithm. In each outer-loop step we reset the bound, i.e., ensure $F_{\text{bound}}(\hat{p}, q, q^{\text{old}}) = F(\hat{p}, q)$. In the inner loop we solve the now convex constrained minimization problem, guaranteeing $F(\hat{p}^{\text{new}}, q^{\text{new}}) \leq F_{\text{bound}}(\hat{p}^{\text{new}}, q^{\text{new}}, q^{\text{old}}) \leq F_{\text{bound}}(\hat{p}, q, q^{\text{old}}) = F(\hat{p}, q)$, while satisfying the constraints (7).

The constrained minimization of (8) in the inner loop can be turned into unconstrained maximization over Lagrange multipliers $\delta_t$ of the functional

$$F_1(\gamma, \delta) = -\sum_{t=1}^{T} \log Z_t \text{ with}$$
$$Z_t = \int d\mathbf{x}_{t-1,t} \, e^{\alpha_{t-1}^T \mathbf{f}(\mathbf{x}_{t-1})} \psi_t(\mathbf{x}_{t-1,t}) e^{\beta_t^T \mathbf{f}(\mathbf{x}_t)} , \quad (9)$$

if we define $\log q^{\text{old}}(\mathbf{x}_t) \equiv \gamma_t \mathbf{f}(\mathbf{x}_t)$ (plus irrelevant constants) and substitute

$$\alpha_t = \frac{1}{2}(\gamma_t + \delta_t) \text{ and } \beta_t = \frac{1}{2}(\gamma_t - \delta_t) . \quad (10)$$

That is, $\delta$ can be interpreted as the difference between the forward and backward messages, $\gamma$ as their sum.

Sketch of proof. Get rid of all terms depending on $q_t(\mathbf{x}_t)$ by substituting (any other convex combination will work as well, but this symmetric one appears most natural)

$$\langle \mathbf{f}(\mathbf{x}_t) \rangle_{q_t} = \frac{1}{2} \left[ \langle \mathbf{f}(\mathbf{x}_t) \rangle_{\hat{p}_t} + \langle \mathbf{f}(\mathbf{x}_t) \rangle_{\hat{p}_{t+1}} \right] .$$

This leaves only the constraints "forward equals backward", $\langle \mathbf{f}(\mathbf{x}_t) \rangle_{\hat{p}_t} = \langle \mathbf{f}(\mathbf{x}_t) \rangle_{\hat{p}_{t+1}}$. The resulting minimization problem in $\hat{p}$ is convex with linear constraints. Introduce Lagrange multipliers $\delta_t$ for these constraints. Minimization of the Lagrangian with respect to $\hat{p}$ yields a distribution of the form (1) if we make the substitutions (10). Plugging this solution back into the Lagrangian yields (9).

The unconstrained maximization problem is concave and has a unique maximum. Any optimization algorithm will do, but a particularly efficient and elegant one is obtained by considering the fixed-point equations. In terms of the standard forward and backward updates $\tilde{\alpha}_t \equiv \tilde{\alpha}_t(\alpha_{t-1}, \beta_t)$ and $\tilde{\beta}_t \equiv \tilde{\beta}_t(\alpha_t, \beta_{t+1})$, the gradient with respect to $\delta_t$ reads

$$\frac{\partial F_1(\gamma, \delta)}{\partial \delta_t} = \frac{1}{2} \left[ \mathbf{g}(\tilde{\alpha}_t + \beta_t) - \mathbf{g}(\alpha_t + \tilde{\beta}_t) \right] . \quad (11)$$

Setting the gradient to zero suggests the update $\delta_t^{\text{new}} = \tilde{\delta}_t \equiv \tilde{\alpha}_t - \tilde{\beta}_t$. This update may be too greedy, but since $\tilde{\delta}_t$ is in the direction of the gradient (11), an increase in $F_1(\delta)$ can be guaranteed with each update

$$\delta_t^{\text{new}} = \delta_t + \epsilon_\delta (\tilde{\delta}_t - \delta_t) , \quad (12)$$

for sufficiently small $\epsilon_\delta$. This update can be loosely interpreted as a natural gradient-ascent step. With each update, one can easily check whether $F_1(\gamma, \delta)$ indeed increases and lower $\epsilon_\delta$ if necessary. In practice, we can often keep $\epsilon_\delta$ at 1.

The outer-loop step can be rewritten as the update

$$\gamma_t^{\text{new}} = \mathbf{g}^{-1} \left( \frac{1}{2} \left[ \mathbf{g}(\alpha_t + \tilde{\beta}_t) + \mathbf{g}(\tilde{\alpha}_t + \beta_t) \right] \right) . \quad (13)$$

### 3.3 SADDLE-POINT PROBLEM

Minimization of the free energy (6) under the constraints (7) is equivalent to the saddle-point problem

$$\min_\gamma \max_\delta F(\gamma, \delta) \text{ with } F(\gamma, \delta) \equiv F_0(\gamma) + F_1(\gamma, \delta),$$
$$\text{and } F_0(\gamma) = \sum_{t=1}^{T-1} \log \int d\mathbf{x}_t \, e^{\gamma_t^T \mathbf{f}(\mathbf{x}_t)} . \quad (14)$$

Sketch of proof. The bound (8), i.e., the outer-loop step in the double loop algorithm, can be written as a minimization over auxiliary variables $\gamma_t$, as e.g. explained in (Minka, 2001a). The maximization over Lagrange multipliers $\delta$ follows when we explicitly write out the inner loop.

The double-loop algorithm basically solves this saddle-point problem (14). Full completion of the maximization in the inner loop is required to guarantee convergence to a correct saddle point. Below we will show that a damped version of the full updates $\alpha_t = \tilde{\alpha}_t$ and $\beta_t = \tilde{\beta}_t$ can be loosely interpreted as a gradient descent-ascent procedure on the same (14). Gradient descent-ascent is a standard approach for finding saddle points of an objective function. Convergence to an, in fact unique, saddle point can be guaranteed if $F(\gamma, \delta)$ is convex in $\gamma$ for all $\delta$ and concave in $\delta$ for all $\gamma$, provided that the step sizes are sufficiently small (Seung et al., 1998). In our case $F(\gamma, \delta)$ is concave in $\delta$, but need not be convex in $\gamma$. The most we can say then is that *stable* fixed points of (damped) expectation propagation must be (local) *minima* of the Bethe free energy (6). The converse need not be the case: minima of the free energy can be unstable fixed points of (damped) expectation propagation.

Sketch of proof. Consider parallel application of damped versions of the inner-loop update (12) and outer-loop update (13). Both updates are aligned with the respective gradients and combining them can therefore be interpreted as performing gradient descent in $\gamma$ and gradient ascent in $\delta$. Choosing $\epsilon_\gamma = 2\epsilon_\delta = 2\epsilon$ and defining $\bar{\gamma}_t \equiv \bar{\alpha}_t + \bar{\beta}_t$ and

$$\Delta_t \equiv 2\mathbf{g}^{-1}\left(\frac{1}{2}\left[\mathbf{g}(\alpha_t + \bar{\beta}_t) + \mathbf{g}(\bar{\alpha}_t + \beta_t)\right]\right)$$
$$- \left([\alpha_t + \bar{\beta}_t] + [\bar{\alpha}_t + \beta_t]\right),$$

we can write the damped update in $\gamma_t$ in the form

$$\gamma_t^{\text{new}} = \gamma_t + \epsilon(\Delta_t + \bar{\gamma}_t - \gamma_t), \quad (15)$$

To study the local stability of this update procedure, we define the Hessian

$$H_{\gamma\gamma} \equiv \left.\frac{\partial^2 F(\gamma, \delta)}{\partial \gamma \partial \gamma^T}\right|_{\gamma^*, \delta^*},$$

and similarly $H_{\delta\gamma}$ and $H_{\delta\delta}$, all evaluated at a fixed point $\{\gamma^*, \delta^*\}$. Gradient descent-ascent is locally stable at $\{\gamma^*, \delta^*\}$ iff $H_{\gamma\gamma}$ is positive definite and $H_{\delta\delta}$ negative definite. The latter is true by construction for all $\gamma$. Consider $F^*(\gamma) \equiv \max_\delta F(\gamma, \delta)$. Its Hessian $H^*_{\gamma\gamma}$ obeys

$$H^*_{\gamma\gamma} = H_{\gamma\gamma} - H_{\gamma\delta} H_{\delta\delta}^{-1} H_{\delta\gamma}. \quad (16)$$

Therefore, if $H_{\gamma\gamma}$ is positive definite (gradient descent-ascent locally stable), then $H^*_{\gamma\gamma}$ as well (local minimum). The opposite need not be true: $H^*_{\gamma\gamma}$ can be positive definite, where $H_{\gamma\gamma}$ is not. An example of this phenomenon is $F(\gamma, \delta) = -\gamma^2 - \delta^2 + 4\gamma\delta$.

Straightforwardly damping the updates $\alpha_t^{\text{new}} = \bar{\alpha}_t$ and $\beta_t^{\text{new}} = \bar{\beta}_t$, we obtain for $\delta_t$ the update (12), but for $\gamma_t$ the update (15) with $\Delta_t = 0$. Since $\Delta_t$ and its gradients with respect to $\delta$ and $\gamma$ vanish at a fixed point, damped expectation propagation has the same local stability properties as the above gradient descent-ascent procedure.

## 4　SIMULATION RESULTS

We tested our algorithms on randomly generated switching linear dynamical systems. Each of the generated instances corresponds to a particular setting of the potentials $\psi_t(\mathbf{x}_{t-1,t})$. We varied $T$ between 3 and 5, the number of switches between 2 and 4, and the dimension of the continuous latent variables and the observations between 2 and 4. Here we will give a phenomenological description of the simulation results.

We focus on the quality of the approximated beliefs $\hat{P}(\mathbf{x}_t|\mathbf{y}_{1:T})$ and compare them with the beliefs that result from the algorithm of (Lauritzen, 1992) based on the strong junction tree, yielding another conditional Gaussian $P(\mathbf{x}_t|\mathbf{y}_{1:T})$. We will refer to the latter as the exact beliefs, although in fact only the probabilities of the switches and the mean and covariance of the conditional distribution given the switches are exact. As a quality measure we consider the Kullback-Leibler divergence $\sum_{t=1}^T \text{KL}(P_t|\hat{P}_t)$.

In most cases undamped expectation propagation works fine and converges within a couple of iterations.

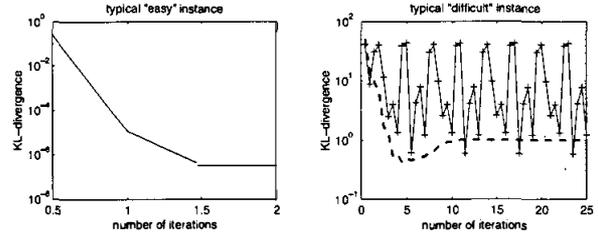

Figure 4: Typical examples of "easy" (left) and "difficult" (right) instances of switching linear dynamical systems with 3 switch states, 3-dimensional continuous latent variables, 4-dimensional observations, and sequence length $T = 5$. The KL-divergence between exact and approximate beliefs is plotted as a function of the number of iterations. Damping with step size $\epsilon = 0.5$ (dashed line) is sufficient to convergence to a stable fixed point.

For a typical instance (see Figure 4 on the left), the KL-divergence drops after a single forward pass (equivalent to GPB2) to an acceptably low value, decreases a little more in the smoothing step, and perhaps a little further in one or two more sweeps until no more significant changes can be seen. Damped approximate belief propagation and the double-loop algorithm converge to the same fixed point, but are less efficient. We will refer to such an instance as "easy".

Occassionally, we ran into a "difficult" instance, where undamped expectation propagation gets stuck in a limit cycle. A typical example is shown in Figure 4 on the right. Here the period of the limit cycle is 8 (eight iterations, each consisting of a forward pass and a backward pass); smaller and even larger periods can be found as well. Damping the belief updates a little, say with $\epsilon = 0.5$ as in Figure 4, is for almost all instances sufficient to converge to a stable solution. The double-loop algorithm always converges as well, with the advantage that no step size has to be set, but usually takes much longer.

We found a single instance in which considerable damping did not lead to convergence. The double-loop algorithm did converge, but the minimum obtained was indeed unstable under single-loop expectation propagation, again even with very small step sizes $\epsilon$. Numerical evaluation of the Hessians at the solution of the double-loop algorithm confirms the analysis around (16) and explains the instability: whereas the Hessian $H^*_{\gamma\gamma}$ of the Bethe free energy $F^*(\gamma) \equiv \max_\delta F(\gamma, \delta)$ is positive definite (local minimum), the Hessian $H_{\gamma\gamma}$ of $F(\gamma, \delta)$ has one significantly negative eigenvalue (gradient descent-ascent unstable).

It has been suggested (see e.g. (Minka, 2001a)) that when undamped (loopy) belief propagation does not converge, it makes no sense to force convergence to the minimum of the Bethe free energy using more heavy artillery: the failure of undamped belief propagation to converge indicates that the solution is inaccurate anyways. To check this hypothesis, we did the following experiment. For each of the "difficult" instances that we found, we generated another "easy" instance with the same structure (length of time sequence, number of switch states, and dimensions). In Figure 5 we have plotted the KL-divergences after a single forward pass (corresponding to GPB2) and after convergence (obtained with damped expectation propagation or the double-loop algorithm for the "difficult" instances). The results suggest the following.

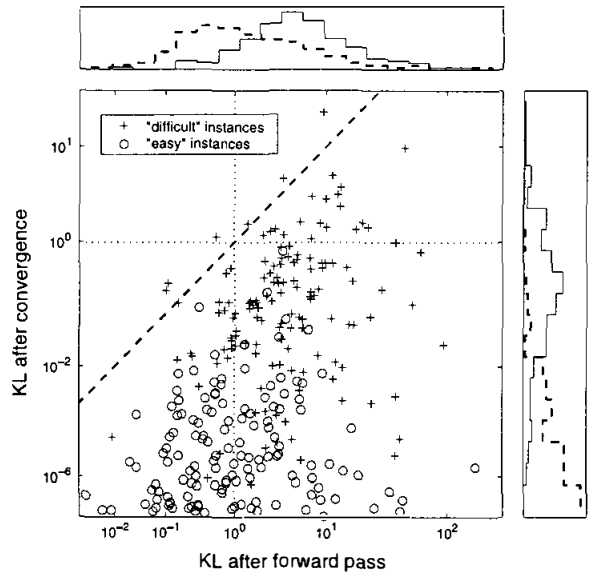

Figure 5: KL-divergences for "easy" ('o') and "difficult" ('+') instances after a single forward pass (GPB2) versus after convergence to a minimum of the free energy. The histograms visualize the distributions of the KL-divergences along the corresponding axes (dashed for "easy" instances, solid for "difficult" ones).

- *It makes sense to iterate and search for the minimum of the free energy*. For almost all instances, both the "easy" and the "difficult" ones, the beliefs corresponding to the minimum of the free energy are closer to the exact beliefs than the ones obtained after a single forward pass.

- *Convergence of undamped belief propagation is an indication, but not a clear-cut criterion for the quality of the approximation.* Indeed, the "easy" instances typically have a smaller KL-divergence than the "difficult" ones. But not always: there is considerable overlap between the KL-divergences for the "easy" and "difficult" instances.

## 5 DISCUSSION AND CONCLUSIONS

We described expectation propagation as a natural extension of exact belief propagation. It has the following crucial ingredients.

1. A description of belief propagation, symmetric with respect to forward and backward messages.

2. The notion to project the beliefs and derive the messages from these approximate beliefs, rather than to approximate the messages themselves.

We derived a convergent double-loop algorithm, similar to the one proposed in (Yuille, 2002) for loopy belief propagation. The bound used here makes it possible to get rid of all $q \log q$-terms, which makes the resulting algorithm slightly more efficient, and, perhaps more importantly, much easier to implement. We interpreted damped expectation propagation as an alternative single-loop algorithm to solve the saddle-point problem (14). It has the nice property that when it converges to a stable fixed point, this must be a minimum of the Bethe free energy. The damped versions suggested in (Minka, 2001a) and (Murphy et al., 1999) for loopy belief propagation are slightly different and may not share this property.

From a practical point of view, undamped expectation propagation works fine in many cases. When it does not, there can still be two different reasons. The innocent reason is a too large step size, similar to taking a too large "learning parameter" in a gradient descent procedure, and is resolved by straightforwardly damping the updates. The more serious reason, which occurred much less frequently in our simulations on switching linear dynamical systems, is when damping does not lead to convergence for very small step sizes. In that case, we can resort to a more tedious double-loop algorithm to guarantee convergence.

Our simulations do not only confirm our theoretical findings, but also suggest that it makes sense to iterate and search for minima of the Bethe free energy, even when undamped expectation propagation fails. Obviously, a more solid comparison would benefit from more simulations on these and different Bayesian networks, comparing them with sampling approaches and other variational techniques. At this point it is very promising that expectation propagation clearly improves upon GPB2 and "solves" the smoothing problem for switching linear dynamical system with hardly any extra implementation efforts. Other issues that deserve more attention are numerical instability (see e.g. (Lauritzen and Jensen, 2001)), as well as the combination of expectation propagation with other (sam-pling) approaches, e.g., when exact computation of the required moments is intractable.

An important question is how the results obtained for chains in this article generalize to general (non-dynamic) Bayesian networks. Preliminary results suggest that one can derive similar double-loop algorithms for guaranteed convergence and single-loop short-cuts with the same correspondence between stable fixed points and local minima. In other words, the results in this article do not seem to be specific to dynamic Bayesian networks, but hold for Bayesian networks in general with projection, loops, or even both. Our current interpretation is that, as soon as messages start to interfere, we have to take care that we update the messages in a special way. For example, going uphill relative to each other to satisfy the constraints, going downhill together to minimize the free energy. That (damped versions of) approximate and loopy belief propagation tend to move in the right uphill/downhill directions might explain why single-loop algorithms converge well in many practical cases.

## Acknowledgements

We would like to thank Taylan Cemgil for helpful input and acknowledge support by the Dutch Technology Foundation STW and the Dutch Centre of Competence Paper and Board.